\pdfoutput=1
\documentclass[11pt]{article}

\usepackage[preprint]{acl}

\usepackage{times}
\usepackage{latexsym}
\usepackage{amsmath}
\usepackage{algorithm}
\usepackage{algorithmic}
\usepackage{booktabs}
\usepackage{multirow}
\usepackage{colortbl}
\usepackage[most]{tcolorbox}
\usepackage{xcolor,soul}

\usepackage{xspace}
\usepackage[T1]{fontenc}
\usepackage[utf8]{inputenc}

\usepackage{microtype}

\usepackage{inconsolata}

\usepackage{graphicx}

\title{Adaptive Token Biaser: Knowledge Editing via Biasing Key Entities}
\author{Baolong Bi$^{1,2}$\quad Shenghua Liu$^{1,2}$\quad  Yiwei Wang$^3$\quad  Lingrui Mei$^{1,2}$ \\ \textbf{Hongcheng Gao}$^2$ \quad \textbf{Yilong Xu}$^{1,2}$ \quad \textbf{Xueqi Cheng}$^{1,2}$ \\
	$^1$CAS Key Laboratory of AI Safety, Institute of Computing Technology, Chinese Academy of Sciences \\
        $^2$University of Chinese Academy of Sciences \\
	$^3$University of California, Los Angeles \quad \\
        \texttt{\small{\{bibaolong23z, liushenghua, xuyilong23s, cxq\}@ict.ac.cn}}   \\ \texttt{ \small{wangyw.evan@gmail.com}}	
        \texttt{\small\{meilingrui22, gaohongcheng23\}@mails.ucas.ac.cn}  
 }

\newcommand{\ours}{\mbox{\textsc{ATBias}}\xspace}
\newcommand{\llamaa}{\mbox{\textsc{LLaMA2-7B-chat}}\xspace}
\newcommand{\llamab}{\mbox{\textsc{LLaMA2-13B-chat}}\xspace}

\newcommand{\Mistral}{\mbox{\textsc{Mistral-7B-instruct}}\xspace}
\newcommand{\CounterFact}{\mbox{\textsc{CounterFact}}\xspace}

\definecolor{ggreen}{rgb}{0.0, 0.6, 0.0}
\definecolor{rred}{rgb}{0.75, 0.0, 0.0}
\definecolor{bblue}{rgb}{0.13, 0.67, 0.8}
\newcommand{\badmetric}[1]{{\color{rred} \textbf{#1}}}
\newcommand{\goodmetric}[1]{{\color{ggreen} \textbf{#1}}}

\definecolor{darkred}{RGB}{200,0,0}
\definecolor{lightgreen}{RGB}{228,253,227}
\definecolor{lightred}{RGB}{252,231,234}
\definecolor{lightyellow}{RGB}{250,253,191}
\definecolor{lightblue}{RGB}{230,240,254}
\definecolor{lightorange}{RGB}{255,223,191}
\definecolor{white}{RGB}{255,255,255}

\newcommand\hlc[2]{\sethlcolor{#1} \hl{#2}}

\newcommand{\orangetext}[1]{{\hlc{lightorange}{#1}}}
\newcommand{\bluetext}[1]{{\hlc{lightblue}{#1}}}

\begin{document}
\maketitle
\begin{abstract}
The parametric knowledge memorized by large language models (LLMs) becomes outdated quickly. 
In-context editing (ICE) is currently the most effective method for updating the knowledge of LLMs. 
Recent advancements involve enhancing ICE by modifying the decoding strategy,  obviating the need for altering internal model structures or adjusting external prompts.
However, this enhancement operates across the entire sequence generation, encompassing a plethora of non-critical tokens.
In this work, we introduce \textbf{A}daptive \textbf{T}oken \textbf{Bias}er (\ours), a new decoding technique designed to enhance ICE.
It focuses on the tokens that are mostly related to knowledge during decoding, biasing their logits by matching key entities related to new and parametric knowledge.
Experimental results show that \ours significantly enhances ICE performance, achieving up to a 32.3\% improvement over state-of-the-art ICE methods while incurring only half the latency.
\ours not only improves the knowledge editing capabilities of ICE but can also be widely applied to LLMs with negligible cost.
\end{abstract}

\section{Introduction}

Large language models (LLMs)~\citep{chatgpt, openai2023gpt4, DBLP:journals/corr/abs-2302-13971, touvron2023llama, song2024fmint} accumulate a substantial volume of factual knowledge during pretraining.
However, some of this knowledge may quickly become outdated, resulting in decreased reliability of LLMs~\citep{chen2023combating, zhang2023sirens, huang2023survey}.
Due to the substantial cost associated with retraining, knowledge editing (KE)~\citep{sinitsin2020editable, de2021editing, mitchell2022memory, yao2023editing} has been proposed to update the knowledge in LLMs by injecting new knowledge or modifying parametric knowledge.

\begin{figure}[t]
    \centering
    \includegraphics[width=\linewidth]{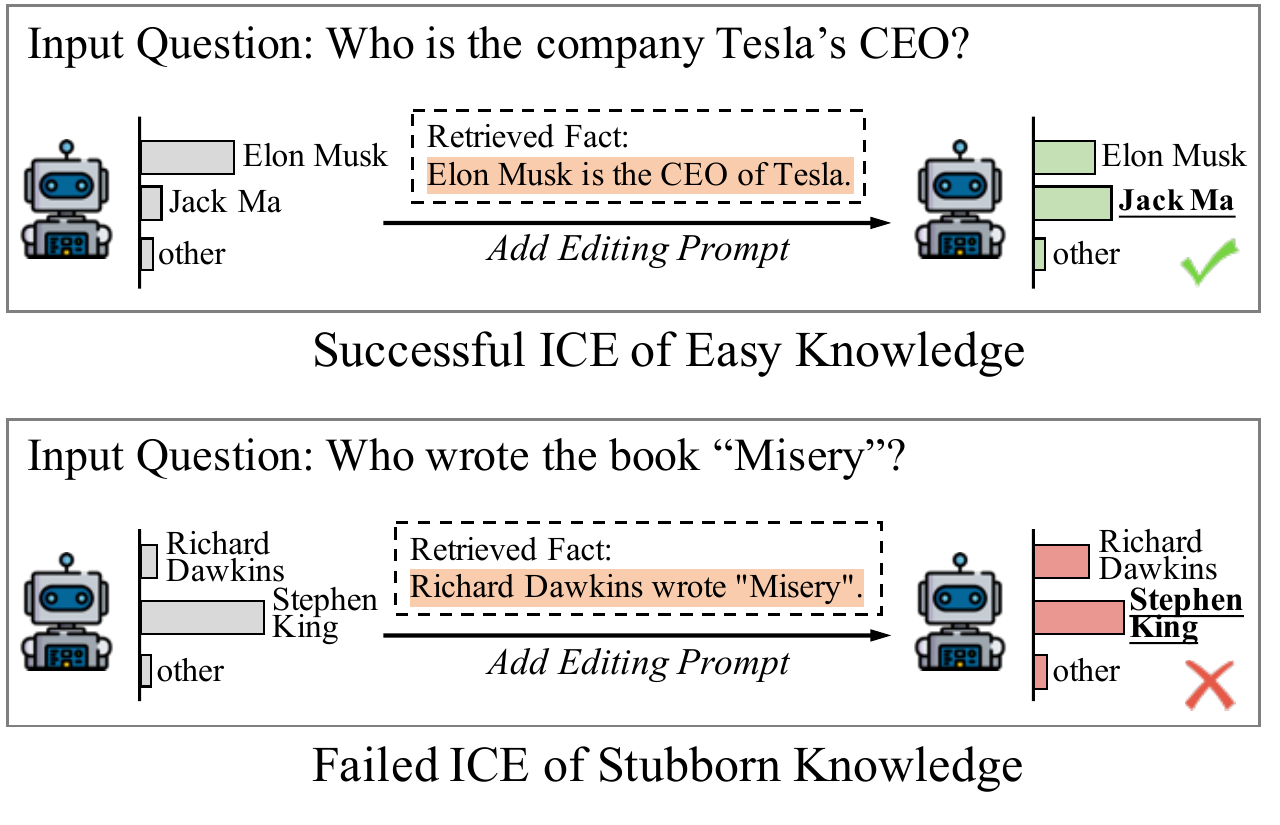}
    \vspace{-8mm}
    \caption{A simple example of in-context editing (ICE). ICE successfully edits easy knowledge but fails to edit stubborn knowledge.}
    \label{fig:showcase}
    \vspace{-6mm}
\end{figure}

As currently the most effective KE method, in-context editing (ICE)~\citep{madaan2022memory, zhong2023mquake, zheng2023can, cohen2024evaluating} has demonstrated state-of-the-art performance in KE.
By providing contextual editing prompts with new knowledge retrieved from the edit memory, ICE can efficiently guide LLMs to inference and generate the answers related to the new knowledge. 

% As currently the state-of-the-art KE method, in-context editing (ICE)~\citep{madaan2022memory, zhong2023mquake, zheng2023can, cohen2024evaluating} provides contextual editing prompts along with new knowledge retrieved from the edit memory, efficiently guiding LLMs in inferring and generating the answers related to the new knowledge.
% wyw: 上面这段不能一个句子。另外上面这个句子没有谓语。
% wyw: most effective -> state-of-the-art

~\citet{bi2024decoding, bi2024factuality} indicate that editing stubborn knowledge solely through external context prompts is challenging, as this knowledge has been established in LLMs with strong confidence during pre-training, as illustrated in Figure \ref{fig:showcase}. 
% wyw: 应该在上一段定义sturbborn knowledge 的 issue
Recent state-of-the-art ICE method DeCK~\citep{bi2024decoding} enhances the editing of stubborn knowledge by modifying entire generating sequence during decoding. 
However, this approach carries potential risks, not only introducing the possibility of inference errors but also incurring higher latency costs.

\begin{figure*}[t]
    \centering
    % \vspace{-8mm}
    \includegraphics[width=\linewidth]{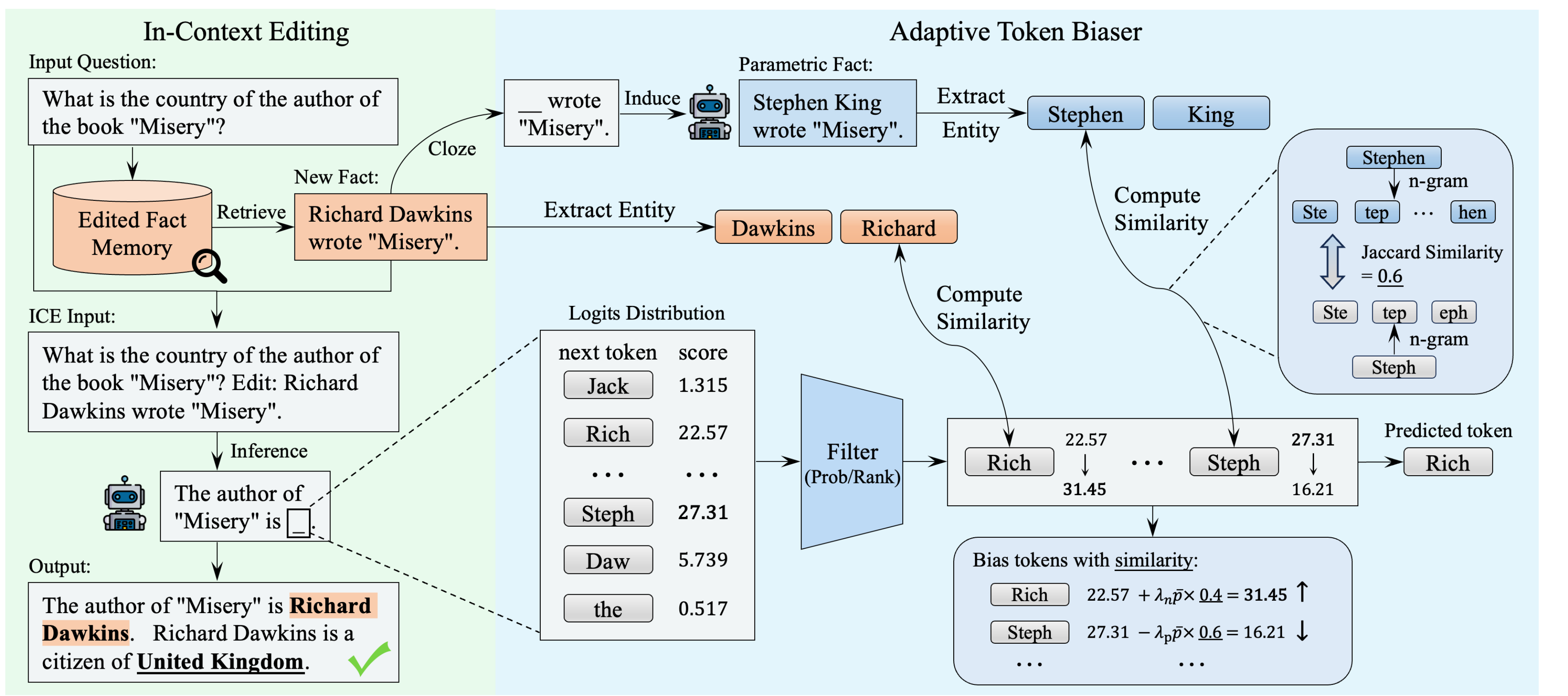}
    \vspace{-5mm}
    \caption{Illustration of how ATBIAS enhances ICE during decoding. \ours adjusts the key token probabilities based on the similarity computed between filtered tokens and extracted new and parametric knowledge entities.}
    \vspace{-3mm}
    \label{fig:framework}
\end{figure*}

In this work, we explore enhancing ICE for editing stubborn knowledge during the decoding stage of LLMs, without altering internal LLMs' parameters or modifying external prompts. 
% wyw: model structures -> LLM parameters
We propose \textbf{A}daptive \textbf{T}oken \textbf{Bias}er (\ours), a new KE framework for LLMs that enhances ICE by matching key entities and biasing the logits of specific tokens. 
The framework of \ours is shown in Figure \ref{fig:framework}.
Unlike previous decoding~\citep{li-etal-2023-contrastive, chuang2023dola, bi2024decoding}, \ours focuses more on the matched tokens rather than the entire generated sequence.
We argue that modifications on other tokens are unnecessary, leading to redundant computational costs and even mistakes.
For example, in generating text \textit{The author Richard Dawkins wrote "Misery"}, the key terms "Richard" and "Dawkins" merit attention over other words in the text. 
Indiscriminate adjustments to other words (such as "The", "author", etc.) can pose a potential risk of introducing fundamental errors in the logical coherence of the entire inference statement.

% The deep insights into ICE reveal that its failure in editing stubborn knowledge is due to the logits of stubborn knowledge being difficult to surpass by the logits of new knowledge.~\citep{bi2024decoding}.
The main goal of \ours is to increase the generation probability of tokens related to new knowledge while decreasing that of parametric knowledge.
Capturing key textual entities is a prerequisite for matching crucial tokens.
\ours provides a parametric induction and entity extraction module, which can efficiently extract key entities from both new facts and parametric facts induced from LLMs.
We also introduce knowledge caching, enabling the aforementioned process to be completed offline. This ensures our \ours performs efficient editing with only a single inference.

% Inspired by adaptive plausibility constraint (APC)~\citep{li-etal-2023-contrastive}, w
We design a specialized filtering mechanism that ensures our approach only considers top-ranked and high-probability predicted tokens. 
The probabilistic-ranking filter not only significantly reduces the likelihood of implausible tokens having their logits erroneously amplified but also greatly improves the time efficiency of \ours.

Tokens related to key entities cannot be precisely located due to the tokenization rules. Therefore, we developed an N-gram and Jaccard-based similarity comparison algorithm to match tokens with entities.
We introduce bias to the logits of both new and old knowledge entities based on the similarity computed between the filtered tokens and these entities.
The tokens related to new knowledge are more likely to be output than parametric knowledge during the generation of LLMs, thus significantly enhancing the editing capabilities of ICE.

Experimental results indicate that our \ours significantly enhances ICE performance, achieving up to a 32.3\% improvement over state-of-the-art decoding methods while incurring only half the latency.
This means that \ours not only further improves editing capabilities but can also be widely applied to LLMs with negligible cost.
Furthermore, we suggest that research into decoding methods should focus more on key tokens rather than the entire sequence  in generation. 
% Our work paves the way for development of both knowledge editing and decoding methods for LLMs.
% wyw: 上面这句话不太对，必须开创某个方向才能用pave the way。

\section{Preliminary}

\paragraph{LLMs Decoding.}
\label{sec:decoding}
The primary goal of LLMs during decoding is to predict the succeeding word within a provided context sequence. Formally, given a sequence of tokens $\{x_1, x_2, \ldots, x_{t-1}\}$ of length $t-1$, we can calculate the probability distribution of next token over the vocabulary set $\mathcal{V}$:
\begin{equation}
     P(x|{x}_{<t}) = \mathrm{softmax}(\phi(h_t)), \quad x \in \mathcal{V}
\label{eq:model_decoding}
\end{equation}
where $\phi(\cdot)$ represents an affine layer for embedding vectors $H=\{h_1, \ldots, h_{t-1}\}$.
In decoding, LLMs samples from the conditional distribution $P(x|{x}_{<t})$ to generate next token $x_t$, continuing this process until an end-of-sequence token is produced. 
% For example, nucleus sampling~\citep{holtzman2019curious} selects tokens from the top-$p$ percentile of the distribution
% % , while top-$k$ sampling ~\citep{fan2018hierarchical} chooses tokens from the top-$k$ candidates.

\paragraph{Multi-hop Editing.}
\label{sec:multi-hop}
Multi-hop editing is a highly challenging task in KE, aimed at verifying whether a fact has been thoroughly edited in LLMs.
It not only edits the specific knowledge but also all related knowledge within the multi-hop relations impacted by this edit.
For example, consider the two-hop question in Figure \ref{fig:framework}. 
The original answer would be "United States" with the facts \textit{Stephen King wrote "Misery", Stephen King is a U.S. citizen}.
With an edit \textit{Richard Dawkins wrote "Misery"} and existing knowledge \textit{Richard Dawkins is British}, the edited output answer should be "United Kingdom".

\section{Methods}

The framework of \ours is shown in Figure \ref{fig:framework}.
First, we induce LLMs to output parametric knowledge by clozing the retrieved new knowledge, and then we extract the knowledge entities from them (Section \ref{sec:preprocess}). This process can be optimized through knowledge caching (Section \ref{sec:cache}). Next, we refine the tokens using a probability and rank-based token filter (Section \ref{sec:filter}), and match key entities with an n-gram and jaccard similarity calculation algorithm (Section \ref{sec:similarity}). Finally, we adaptively bias the logits of the crucial tokens (Section \ref{sec:adjustment}) to predict the next tokens.

\vspace{2mm}

\subsection{Parametric Induction \& Entity Extraction}
\label{sec:preprocess}
Extracting key knowledge entities from redundant knowledge information is a fundamental prerequisite of \ours.  
This enables the adjustment of corresponding token probabilities during decoding.
Specifically, \ours enables the preprocessing to obtain parametric output from LLMs corresponding to each new fact piece in the edit memory. 
For example, consider a piece of new fact updated in the edited fact memory: \textit{The author Richard Dawkins wrote "Misery"}. 
By clozing the new fact such as \textit{The author \_ wrote "Misery"}, LLMs can be induced to provide parametric fact outputs like \textit{The author Stephen King wrote "Misery"}. 
% It's worth noting that this operation is performed offline rather than during online inference stages.

Subsequently, the key knowledge entities are individually extracted from these fact pieces. 
We define the function $\text{extract}(\cdot)$ to represent this process. Given a set of fact pieces $fact$, we can obtain a list of split entity strings:

\begin{equation}
    E_{fact}=\text{Extract}(fact)
\end{equation}
Then, the extracted entities $E_{\text{new}}$ and $E_{\text{para}}$ from new fact and parametric fact are used to match the key tokens in Section \ref{sec:adjustment}.

% Actually, offline preprocessing is not imperative, as many advanced ICE methods~\citep{zhong2023mquake, wang2024deepedit, shi2024retrieval} inherently involve parametric output during their question-answering process with LLMs.
% For example, MeLLo~\citep{zhong2023mquake} prompts LLMs to output parametric answers to subquestions and checks for conflicts with retrieved facts.
% Therefore, \ours can also extract aligned entities online with negligible time overhead, using simple methods or tools such as fine-tuned LMs or regular expressions.

% \vspace{2mm}
\subsection{Probabilistic-Ranking Filter}
% \vspace{2mm}

\label{sec:filter}
As introduced in Section \ref{sec:decoding}, tokens with higher probabilities in the distribution $P(x|{x}_{<t})$ are more likely to be sampled and output during the decoding in LLMs.
However, if we calculate the similarity (Section \ref{sec:similarity}) for all tokens in the vocabulary $\mathcal{V}$ to adjust their logits (Section \ref{sec:adjustment}), it will not only cause unnecessary time overhead but also increase the potential risk of erroneously amplifying the probabilities of unreliable tokens.

Inspired by APC~\citep{li-etal-2023-contrastive}, we design a stringent filtering mechanism to eliminate the unreliable tokens.
Specifically, we control the decoding scope based on both the probability values of the tokens and their rankings.

First, we set a constraint parameter $\alpha$ to ensure that the filtered tokens logits have only a small difference from the highest probability.
Using $P(x_t)$ to represent $P(x_t|x_{<t})$ for notational brevity,
the probabilistic filter can be formalized as follows:
\begin{equation}
\mathcal{V}_\text{prob}= \left\{x_t \in \mathcal{V}: P(x_t) \geq \alpha \max _w P{(w)}\right\}
\label{eq:prob}
\end{equation}
We then define the ranking-based filtering:
\begin{equation}
\mathcal{V}_\text{rank}= \left\{x_t \in \mathcal{V}: P(x_t) \geq P(R_k)\right\}
\label{eq:rank}
\end{equation}
where $R_k$ represents the token with the k-th largest probability. 
This implies that we exclusively focus on the top-k tokens in the distribution. Subsequently, we obtain a more stringent set of filtered tokens:
\begin{equation}
\mathcal{V}_{\text {head}}(x_t|x_{<t})= \mathcal{V}_1 \cap \mathcal{V}_2
\label{eq
}
\end{equation}
$\mathcal{V}_{\text {head}}$ imposes specific decoding constraints by considering both probability and ranking, thereby avoiding situations where filtered tokens have high rankings but low credibility, or where there are too many tokens with high probabilities.
We can then predict the next token by:
\begin{equation}
P_\text{filt}(x_t) = \begin{cases} P(x_t), & \text { if } x_t \in \mathcal{V}_{\text {head}}(x_t|x_{<t}), \\
-\infty , & \text { otherwise. }\end{cases} 
\label{eq:F}
\end{equation}

\subsection{N-gram and Jaccard Similarity}
\label{sec:similarity}
\vspace{2mm}

Tokens related to key entities cannot be precisely identified due to tokenization rules.
Therefore, directly identifying specific tokens and adjusting their logits is not feasible.
\ours presents a novel approach where tokens are first decoded into strings during the decoding process, which are then matched with the strings derived from key entities. 
Thus, tokens more relevant to key entities can be identified by matching decoded strings with entity strings.
An additional challenge is that a word may be segmented by the tokenizer into various prefixes, infixes, and suffixes.  
For example, 'Dawkins' might be segmented into "Daw-" and "-kins".  
This means the decoded strings may be the substrings of entity strings.
Therefore, we need to match the substrings obtained from decoding the tokens with the full strings split from the key entities.

To tackle the above
challenges, we developed an N-gram and Jaccard-based similarity comparison algorithm to match the filtered tokens with key entities.
The target of our algorithm is to compare the similarity between substrings (tokens) and full strings (entities), including complex word structures such as prefixes, infixes, and suffixes.

We begins by decomposing both the substring and the full string into character n-grams. 
A character n-gram is a contiguous sequence of n characters within a string.
We define the function $g(\cdot)$ to represent the decomposition of a string into an n-gram set.
For example, for the string "Stephen" and n=3, the set of 3-grams includes $g(\text{"Stephen"})=\{\text{"Ste"}, \text{"tep"}, \text{"eph"}, \text{"phe"}, \text{"hen"}\}$. 
Specially, we compute the n-gram sets of the substrings and the full strings using a sliding window approach.

Next, we can calculate the Jaccard similarity~\citep{niwattanakul2013using} between the two decomposed n-gram sets of decoded strings and entity strings, which can be formalized as follows:

\begin{equation}
    \text{sim}(x_t, e_i) = \frac{|g(x^\text{d}_t) \cap g({e_i})|}{|g(x^\text{d}_t) \cup g({e_i})|}
\end{equation}
where $x^\text{d}_t$ represents the decoded strings from filtered tokens satisfying $x^\text{d}_t=\text{decode}(x_t)$ and $x_t \in \mathcal{V}_{\text {head}}(x_t|x_{<t})$,  $e_i \in E$ and $E=\{e_1, e_2, \ldots, e_m\}$ is the split entity strings set of length $m$.

\begin{algorithm}[h]
\caption{Adaptive Token Biaser}
\label{alg:ours}
\begin{algorithmic}[1]
\REQUIRE $P$: distribution of tokens, $\mathcal{V}$: vocabulary, $\mathcal{F}_\text{new}$: new facts, $\mathcal{F}_\text{para}$: parametric facts
\STATE Filter $\mathcal{V} \rightarrow \mathcal{V}_{\text{head}}$
\STATE Extract $\mathcal{F}_\text{new} \rightarrow E_{\text{new}}$ and $\mathcal{F}_\text{para} \rightarrow E_{\text{para}}$
\STATE Compute avg. $\mathrm{\bar{P}} = \frac{1}{|\mathcal{V}_{\text{head}}|} \sum_{x_i \in \mathcal{V}_{\text{head}}} P(x_i)$
\FOR{each $x_i \in \mathcal{V}_{\text{head}}$}
    \STATE Decode $x_i \rightarrow x_i^\text{d}$
    \FOR{each $e_j^\text{n} \in E_{\text{new}}$}
        \IF{$x_i^\text{d}$ in $e_j^\text{n}$}
            \STATE N-gram Decompose $x_i^\text{d}$, $e_j^\text{n}$ $\rightarrow$ $g_x^i$, $g_e^j$
            \STATE $P(x_i) = P(x_i) + \lambda_{\text{n}} \cdot \mathrm{\bar{P}} \cdot  \frac{|g_x^i \cap g_e^j|}{|g_x^i \cup g_e^j|}$
        \ENDIF
    \ENDFOR
    \FOR{each $e_j^\text{p} \in E_{\text{para}}$}
        \IF{$x_i^\text{d}$ in $e_j^\text{p}$}
            \STATE N-gram Decompose $x_i^\text{d}$, $e_j^\text{p}$ $\rightarrow$ $g_x^i$, $g_e^j$
            \STATE $P(x_i) = P(x_i) - \lambda_{\text{p}} \cdot \mathrm{\bar{P}} \cdot  \frac{|g_x^i \cap g_e^j|}{|g_x^i \cup g_e^j|}$
        \ENDIF
    \ENDFOR
\ENDFOR
\RETURN $P$
\end{algorithmic}
\end{algorithm}

\subsection{Adaptive Token Biaser}
\label{sec:adjustment}
% \vspace{1mm}
The main goal of our adaptive token biaser is to increase the logits of tokens corresponding to new knowledge entities while decreasing those of parametric knowledge entities, thus enhancing the capability of ICE editing knowledge.
Therefore, the biasing operation can be divided into two parts, starting with the enhancement of new knowledge:
\begin{equation}
     P_{\text{adj}}(x_t) = P_{\text{filt}}(x_t) + \lambda_\text{n} \mathrm{\bar{P}} \text{sim}(x_t,e^{\text{n}}_i) 
\label{eq:adjustment1}
\end{equation}
where $\lambda_\text{n}$ is the bias coefficient for new knowledge, $e^{\text{n}}_i$ represents the split string of new knowledge entities such that $e^{\text{n}}_i \in E_{\text{new}} = \{e^{\text{n}}_1, e^{\text{n}}_2, \ldots, e^{\text{n}}_m\}$. 
And $\mathrm{\bar{P}}$ is the average probability of all filtered tokens, defined as:
\begin{equation}
\mathrm{\bar{P}} = \frac{1}{|\mathcal{V}_{\text{head}}|} \sum_{x_t \in \mathcal{V}_{\text{head}}} P_{\text{filt}}(x_t)
\end{equation}
Similarly, the suppression of parametric knowledge can be represented as follows:
\begin{equation}
     P_{\text{adj}}(x_t) = P_{\text{filt}}(x_t) - \lambda_\text{p} \mathrm{\bar{P}} \text{sim}(x_t,e^{\text{p}}_i) 
    \label{eq:adjustment2}
\end{equation}
where $\lambda_\text{p}$ is the tuning coefficient for parametric knowledge, $e^{\text{p}}_i \in E_{\text{para}}$ represents the split string of parametric knowledge entities.

\begin{table*}[t!]
\centering
\small
\renewcommand{\arraystretch}{1}
\resizebox{\linewidth}{!}{
\begin{tabular}{llccc}
\toprule
\textbf{Model} & {\textbf{Method}} &  {\textbf{\textsc{MQuAKE-3k}}} &{\textbf{\textsc{MQuAKE-2002}}} & {\textbf{\textsc{MQuAKE-hard}}}\\
\midrule
& ROME~\citep{meng2022locating} & 18.2 & 19.1 & 15.7 \\
 {\textsc{LLaMA2-}}& IKE~\citep{zheng2023can} & 85.4 & 85.1 & 88.9 \\
{\textsc{7B-chat}} & IKE w/ DeCK~\citep{bi2024decoding} & 91.3 & 89.4 & 98.6 \\
& IKE w/ \ours (ours) & \bf 93.1 & \bf 92.3 & \bf 98.8 \\
\midrule
 & ROME~\citep{meng2022locating} & 39.4 & 39.7 & 35.2\\
{\textsc{LLaMA2-}} & IKE~\citep{zheng2023can} & 63.8 & 64.1 & 55.2\\
{\textsc{13B-chat}} & IKE w/ DeCK~\citep{bi2024decoding}  & 84.6 & 84.4 & 89.7\\
& IKE w/ \ours (ours) & \bf 89.7 & \bf 87.6 & \bf 91.2 \\
\midrule
 & ROME~\citep{meng2022locating} & 28.1 & 30.2 & \bf 26.3\\
  \textsc{Mistral-} & IKE~\citep{zheng2023can} &  34.1 &  35.6 &  15.6\\
  {\textsc{7B-instruct}} & IKE w/ DeCK~\citep{bi2024decoding}  & 46.7 & 48.5 & 19.2\\
  & IKE w/ \ours (ours) & \bf 47.6 & \bf 48.7 & 22.6 \\
\bottomrule
\end{tabular}
}
% \vspace{8pt}
\caption{Experimental results (accuracy; \%) across ROME, original IKE, IKE enhanced by DeCK and our \ours. The batch size of the edit memory was set to 1 to evaluate the foundational capability of directly editing knowledge. The best editing result for each LLM is highlighted in bold font.}
\vspace{-12pt}
\label{tab:direct}
\end{table*}

The overall process of \ours is shown in Algorithm \ref{alg:ours}.
\ours controls the degree of logits bias for tokens corresponding to key entity texts by calculating similarity. 
The n-gram and jaccard similarity in Section \ref{sec:similarity} ensures the validity of this step, as a higher overlap receives a certain weight, while lower overlap or mismatches receive a weight of zero.
This distinguishes our \ours from the decoding methods that operate on the entire generating sequence.
\ours focuses only on the few crucial tokens, such as "Richard" and "Dawkins" in \textit{The author Richard Dawkins wrote "Misery"}, without biasing the majority of others like "The", "author", etc.
This ensures that our editing process does not interfere with the reasoning of LLMs, reducing the potential risk of introducing inappropriate tokens during decoding.

\subsection{Knowledge Caching for Efficient Editing}
\label{sec:cache}
Considering that parametric induction and entity extraction in Section \ref{sec:preprocess} can introduce additional time overhead, we can preprocess these steps in advance.  Specifically, whenever a new fact is updated in the edited memory, we offline induce the LLMs to output the corresponding parametric fact and then extract the entities from both the new and parametric facts.  We record these in a knowledge cache to ensure that they can be directly retrieved during online inference by the LLMs.

Actually, the offline preprocessing is not imperative, as many advanced ICE methods~\citep{zhong2023mquake, wang2024deepedit, shi2024retrieval} inherently involve parametric output during their process with LLMs.
For example, MeLLo~\citep{zhong2023mquake} prompts LLMs to output parametric answers to subquestions. 
And then \ours can extract the entities from these parametric answers in MeLLo online, using simple methods or tools such as fine-tuned LMs or regular expressions. 
See the Appendix \ref{sec:mello_example} for detailed examples.
Therefore, our \ours only requires a single inference with negligible additional overhead.

\section{Experiments}

\subsection{Experimental Setup}

\paragraph{Tasks.}
Our experiments focus on the one-hop and multi-hop question-answering tasks introduced in Section \ref{sec:multi-hop}.
We set the batch size of the edit memory as 1 and full batch for multi-hop editing evaluation.
The batch size means the number of instances providing the edited facts for knowledge retrieval.

\vspace{-4pt}

\paragraph{Datasets.}
We conduct extensive experiments for the main multi-hop editing task using \textsc{MQuAKE-3k}~\citep{zhong2023mquake} along with its challenging derivatives, \textsc{MQuAKE-2002} and \textsc{MQuAKE-hard}, introduced by~\citet{wang2024deepedit}. 
\textsc{MQuAKE} provides multi-hop knowledge questions to evaluate KE on counterfactual edits.
We also evaluate for one-hop editing task on \textsc{CounterFact}~\citep{meng2022locating}.
Additionally, we follow~\citep{bi2024decoding} to use corresponding \textsc{stubborn} datasets to further evaluate the effectiveness of editing stubborn knowledge in Section~\ref{sec:stubborn}.

\begin{table*}[t!]
\centering
\small
\renewcommand{\arraystretch}{1}
\resizebox{\linewidth}{!}{
\begin{tabular}{llccc}
\toprule
\textbf{Model} & {\textbf{Method}} &  {\textbf{\textsc{MQuAKE-3k}}} &{\textbf{\textsc{MQuAKE-2002}}} & {\textbf{\textsc{MQuAKE-hard}}}\\
\midrule
   \multirow{2}{*}{\textsc{LLaMA2-}} & MeLLo~\citep{zhong2023mquake} & 32.6 & 40.8 & 5.1\\
   \multirow{2}{*}{\textsc{7B-chat}} & MeLLo w/ DeCK~\citep{bi2024decoding} & 43.1 & 45.8 & 5.8\\
   & MeLLo w/ \ours (ours) & \bf 54.3 & \bf 48.9 & \bf 6.3\\
\midrule
   \multirow{2}{*}{\textsc{LLaMA2-}} & MeLLo~\citep{zhong2023mquake} & 33.4 & 35.9 & 3.9\\
   \multirow{2}{*}{\textsc{13B-chat}} & MeLLo w/ DeCK~\citep{bi2024decoding}  & 36.8 & 38.2 & 6.2\\
   & MeLLo w/ \ours (ours) & \bf 48.7 & \bf 43.6 & \bf 6.7\\
\midrule
   \multirow{2}{*}{\textsc{Mistral-}} & MeLLo~\citep{zhong2023mquake} & 21.8 & 22.8 & 2.1\\
   \multirow{2}{*}{\textsc{7B-instruct}} & MeLLo w/ DeCK~\citep{bi2024decoding}  & 21.3 & 22.9 & 2.6\\
   & MeLLo w/ \ours (ours) & \bf 24.7 & \bf 25.4 & \bf 3.1\\
\bottomrule
\end{tabular}
}
% \vspace{8pt}
\caption{Experimental results (accuracy; \%) on multi-
hop editing task with 500 instances. We conduct the experiments with the full batch size edit memory to evaluate the performance of memory based KE.}
\vspace{-8pt}
\label{tab:retrieval}
\end{table*}

\vspace{-4pt}
\paragraph{Models and Baselines.}
We examine different LLM families and sizes, including \textsc{LLaMA2-7B-chat}, \textsc{LLaMA2-13B-chat}\citep{touvron2023llama}, and \Mistral\citep{jiang2023mistral}.
We employ the state-of-art ICE methods IKE~\citep{cohen2024evaluating} and MeLLo~\citep{zhong2023mquake}, and advanced model-editing techniques ROME~\citep{meng2022locating} as baselines.
We also compare our approach with these ICE methods enhanced by DeCK~\cite{bi2024decoding}, the state-of-the-art decoding method for ICE that contrasts knowledge.
IKE prompts LLMs to edit new knowledge using contextual demonstrations, while MeLLo edits multi-hop knowledge by decomposing sub-questions and guiding LLMs to generate answers. ROME views editing as least squares with linear equality constraints and employs the Lagrange multiplier for solving.

\vspace{-4pt}

\paragraph{Implementation.}
% Our approach \ours does not require altering internal model structures or adjusting external prompts. 
% It operates solely within the decoding process of the transformers architecture, making it easily deployable onto the ICE methods IKE and MeLLo.
We implement IKE with multi-hop question-answering demonstrations and chain-of-thought (COT) ~\citep{wei2022chain} prompting to enhance its performance. 
We deploy \ours to MeLLo without the need for additional preprocessing, as MeLLo naturally outputs parametric knowledge (Section~\ref{sec:preprocess}).
The prompts used in IKE and MeLLo are shown in Appendix \ref{sec:prompts}.
The model editing methods ROME in our baselines are deployed using EasyEdit~\citep{wang2023easyedit}. 
We set n to 2 in the n-gram decomposition, the adaptive constraint $\alpha$ to 0.0005 and $k$ to 10, with bias coefficients $\lambda_{n}$ set to 25 and $\lambda_{p}$ set to 1.

\subsection{Overall Performance}

We set the batch size of the edit memory as 1 for evaluating the foundational direct editing capabilities of IKE~\citep{zheng2023can} method, especially considering multi-hop questions with 1,000 instances.
The batch size means the number of instances providing the edited facts for knowledge retrieval.
Table \ref{tab:direct} displays the performance on \textsc{MQuAKE} across various models and datasets.
Overall, compared to the model-editing method ROME, the ICE method IKE demonstrates a clear advantage.  
The enhanced IKE by \ours consistently shows the best performance.
Furthermore, as model parameters increase (\llamab) and pretraining becomes more refined (\Mistral), the knowledge within LLMs becomes more stubborn to editing.  
\ours can enhance ICE to effectively edit this stubborn knowledge.

We follow the setup of previous work \citep{zhong2023mquake, wang2024deepedit} to conduct experiments for MeLLo~\citep{zhong2023mquake} with the full batch size edit memory.
As shown in Table \ref{tab:retrieval}, the experimental results illustrate that \ours enhances MeLLo to varying degrees in full batch experiments.
Specifically, the enhancement provided by our \ours shows a significant advantage, with an impressive improvement of up to 32.3\% compared to DeCK.
This is because \ours operates on a small number of key tokens rather than the entire sequence as in DeCK, leaving other tokens in the inference process unaffected.  
This greatly reduces the potential risk of introducing fundamental errors during the inference stage, making our \ours's enhancements even more pronounced in longer and more complex editing pipelines.
It further indicates that \ours holds significant potential for real-world KE applications with higher performance and lower costs.

\subsection{One-hop Editing}
\label{sec:one-hop}
Despite the greater challenge of multihop editing, we still used the \CounterFact dataset to evaluate one-hop editing for the robustness of our method. 
As the results shown in Table~\ref{tab:one-hop}, the ICE method IKE achieved high accuracy in the simpler one-hop editing task, with IKE enhanced by our \ours consistently outperforming others.

\vspace{2pt}
\begin{table}[ht]
\centering
\small
\renewcommand\arraystretch{1.3}
\setlength{\tabcolsep}{6pt}
\begin{tabular}{lccc} 
\toprule
 \ \textbf{Model} & \textbf{IKE}  & \textbf{w/ DeCK} & \textbf{w/ \ours}\\ \midrule 
 \textsc{LLaMA2-7B} & 98.37 & 98.65 & \bf 99.42 \\ 
 \textsc{LLaMA2-13B} & 93.76 & 94.23 & \bf 95.35 \\ 

\bottomrule
\end{tabular}
\caption{Experimental results of IKE on \textsc{CounterFact} for one-hop editing task.} 
\label{tab:one-hop} 
\vspace{-10pt}
\end{table}

\begin{table*}[ht]
\centering
% \small
\renewcommand{\arraystretch}{1}
\resizebox{0.9\linewidth}{!}{
\begin{tabular}{llcccc}
\toprule
\textbf{Model} & {\textbf{\textsc{stubborn}}} &{\textbf{ROME}} & {\textbf{IKE}} & {\textbf{IKE w/ DeCK}} & {\textbf{IKE w/ \ours}}\\
\midrule
 \multirow{2}{*}{\llamaa} & $>33\%$ & 17.7 & 56.4 & 72.3 & \bf73.9\\
 & $>67\%$ & 19.3 & 37.8 & 55.9 & \bf57.8\\
\midrule
  \multirow{2}{*}{\llamab}& $>33\%$ & 42.5 & 38.9 & 70.1 & \bf71.6\\
  & $>67\%$ & 40.2 & 29.4 & 48.5 & \bf 56.5\\
 \midrule
  \multirow{2}{*}{\Mistral}& $>33\%$ & 19.7 & 20.7 & 26.5 & \bf 33.2\\
  & $>67\%$ & 18.5 & 17.9 & 22.6 & \bf 27.9\\
  \bottomrule
\end{tabular}
}
% \vspace{-8pt}
\caption{Performance of different models on their respective \textsc{stubborn} datasets. The edit memory batch size of the IKE methods is set to 1. `\textsc{stubborn} > 33\%' indicates instances from the \textsc{MQuAKE-3k} dataset where IKE failed to edit knowledge more than 33\% of the time. `\textsc{stubborn} > 67\%' follows the same criterion.}
\vspace{-8pt}
\label{tab:stubborn}
\end{table*}

\subsection{Editing on Stubborn Knowledge}
\label{sec:stubborn}
Stubborn knowledge in LLMs is difficult to edit because it is established with strong confidence during the pretraining process. 
We follow ~\citep{bi2024decoding} to construct the corresponding \textsc{stubborn} datasets for different models to specifically evaluate \ours's performance on stubborn knowledge. 
The stubborn datasets are categorized into different difficulty levels based on the proportion of correct answers when using ICE methods to edit the same knowledge multiple times with different questions.

The experimental results on the \textsc{stubborn} datasets are presented in Table \ref{tab:stubborn}.
We find that IKE's performance on \textsc{stubborn} datasets significantly declined compared to Table \ref{tab:direct}, even falling below the model-editing method ROME. 
This indicates that relying solely on external prompts is insufficient to change LLMs' confidence in this stubborn knowledge.
The enhancement methods applied during decoding significantly improve the effectiveness of editing stubborn knowledge, with \ours consistently achieving the best performance.  
This suggests \ours enhances ICE methods' ability to effectively edit stubborn knowledge.

\begin{table}[h]
\centering
% \small
\renewcommand{\arraystretch}{1}
\resizebox{\linewidth}{!}{
\begin{tabular}{llcccc}
\toprule
\multirow{2}{*}{\textbf{Model}} & \multirow{2}{*}{\textbf{Method}} & \textbf{Latency} & \textbf{Throughput} \\
 & & \textbf{(ms/token)} & \textbf{(token/s)} \\
\midrule
\multirow{2}{*}{\textsc{LLaMA2-}} & Baseline &  36.03 {\tiny \bf ($\times$1.00)} &  27.76 {\tiny \bf ($\times$1.00)} \\
 \multirow{2}{*}{\textsc{7B-chat}} & DeCK & \badmetric{69.99 {\tiny \bf ($\times$1.94)}} & \badmetric{14.29 {\tiny \bf ($\times$0.51)}} \\
 & \ours & \goodmetric{36.19 {\tiny \bf ($\times$1.01)}} & \goodmetric{27.64 {\tiny \bf ($\times$1.00)}} \\
 \midrule
\multirow{2}{*}{\textsc{LLaMA2-}} & Baseline & {51.41} {\tiny \bf ($\times$1.00)} & {19.45} {\tiny \bf ($\times$1.00)} \\
 \multirow{2}{*}{\textsc{13B-chat}} & DeCK & \badmetric{94.08 {\tiny \bf ($\times$1.83)}} & \badmetric{10.63 {\tiny \bf ($\times$0.55)}} \\
 & \ours & \goodmetric{49.11 {\tiny \bf ($\times$0.95)}} & \goodmetric{20.36 {\tiny \bf ($\times$1.05)}} \\
\bottomrule
\end{tabular}
}
\vspace{-5pt}
\caption{Decoding latency (ms/tokens) and throughput (tokens/s). \goodmetric{Green} shows low latency and high throughput, \badmetric{red} shows high latency and low throughput.}
\vspace{-8pt}
\label{tab:latency}
\arrayrulecolor{black}
\end{table}

\subsection{Latency \& Throughput}
\label{sec:latency}
Table \ref{tab:latency} shows the decoding latency for the baseline, as well as when incorporating DeCK or \ours.
DeCK requires generating and comparing two sequences during decoding, resulting in approximately 2x the latency of the baseline.
It is worth noting that \ours increases the decoding time by only a factor of 1.01 in \textsc{LLaMA2-7B-chat} and is even more efficient in \textsc{LLaMA2-13B-chat} compared to the baseline.
This efficiency is due to the probabilistic-ranking filter, which filters out most low-probability tokens and only considers highly confident tokens for prediction.
It suggests that \ours, with its outstanding editing performance, can also be widely applied at negligible cost.

\begin{figure*}[t!]
    \centering
    % \vspace{-5mm}
    \includegraphics[width=\linewidth]{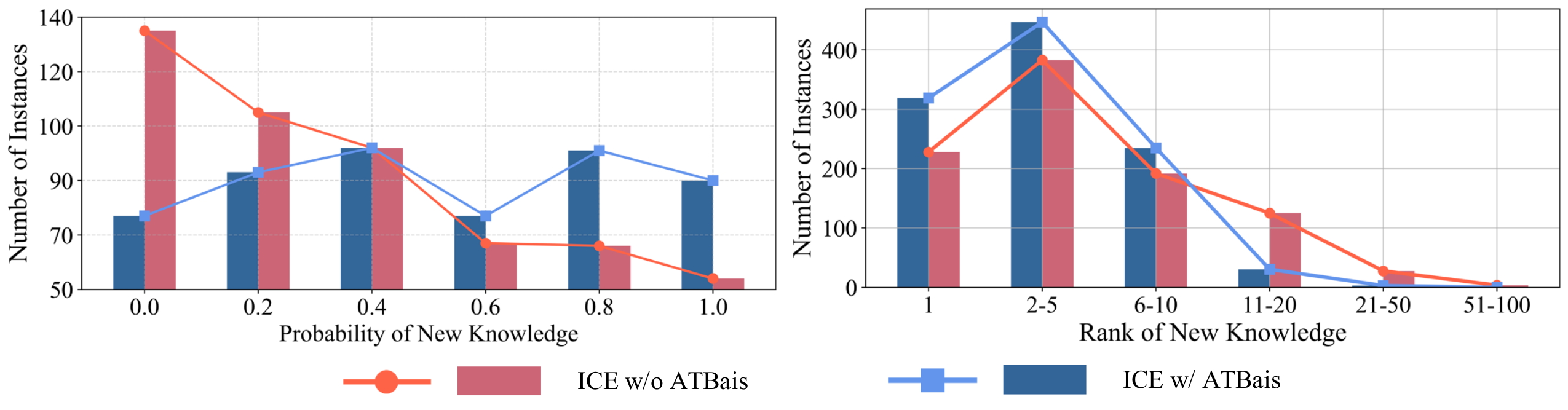}
    \vspace{-6mm}
    \caption{Probability (left) and ranking (right) statistics of new Knowledge for \llamaa on stubborn > 33\%. The probabilities are derived from normalize calculations.}
    \vspace{-2mm}
    \label{fig:stubborn}
\end{figure*}

\subsection{Why \ours Edits Efficiently?}
\label{sec:why}
 
\citet{bi2024decoding} observes that the values of the logits corresponding to the parametric knowledge's tokens are very high before editing. 
Even though ICE significantly increases the logits of the tokens corresponding to new knowledge, there are still cases where it fails to surpass the parametric ones.
To reveal the underlying reasons why \ours can effectively enhance the ICE methods from a model interpretability perspective, we analyzed the probability changes of the new knowledge before and after applying \ours.
Specifically, we capture the first tokens of the new and parametric knowledge entities that represent them and then record their normalized logits.

The results illustrated in Figure \ref{fig:stubborn} show that IKE with \ours has a higher distribution within the high probability range, while IKE without \ours is concentrated in the low probability range.
Additionally, the probability ranking of new knowledge significantly increased after incorporating \ours.
Moreover, the probability distribution of the parametric knowledge exhibited an opposite trend after incorporating \ours.
This further explains why \ours can effectively enhance ICE: it increases the probabilities of new knowledge entities and decreases the probabilities of parametric knowledge entities.
As shown in the editing example in Figure \ref{fig:framework} (\textit{Richard Dawkins is a citizen of the United Kingdom}), the newly generated knowledge entities by \ours serve as new contextual cues during inferencing to reason over multiple hops of knowledge, significantly improving editing performance.

\subsection{Ablation Study}
We conducted a comprehensive ablation study on the adaptive constraints, bias coefficients, and key components of \ours. 
Table \ref{tab:ablation-fileter} presents the results for the filter in \ours, demonstrating the necessity of filtering tokens based on both probability and ranking constraints. 
Additional ablation study results can be found in the Appendix \ref{sec:add_ablation}.

\begin{table}[ht]
\centering
\small
\renewcommand\arraystretch{1.3}
\setlength{\tabcolsep}{8pt}
\begin{tabular}{lccc} 
\toprule
 \ \textbf{Model} & \textbf{Prob}  & \textbf{Rank} & \textbf{Prob \& Rank}\\ \midrule 
 \textsc{LLaMA2-7B} & 90.2 & 81.5 & \bf 93.1 \\ 
 \textsc{LLaMA2-13B} & 81.9 & 72.4 & \bf 89.7 \\ 

\bottomrule
\end{tabular}
\caption{Ablation study results for the filter of our \ours. Prob and Rank respectively represent probability and ranking constraints in the filter.} 
\label{tab:ablation-fileter} 
\vspace{-10pt}
\end{table}

\section{Related Work}
\paragraph{Factual Hallucinations.} 
Factual hallucinations have garnered widespread attention due to their significant side effects, as LLMs generate content that deviates from established world knowledge~\citep{tonmoy2024comprehensive, huang2023survey, wang2023survey, jiang2024hummer}. 
These hallucinations can arise from various sources and at different stages of the LLM life cycle~\citep{zhang2023sirens}.
Outdated knowledge is a major factor contributing to factual hallucinations.
\ours enhances KE during the inference stage in LLMs to mitigate these hallucinations.

\paragraph{Knowledge Editing.} 
KE~\citep{yao2023editing} has been proposed to update information in LLMs, enabling accurate responses to current questions.
In general, there are three lines of works for KE. 
Model editing~\citep{zhu2020modifying, meng2022locating, meng2022mass, huang2023transformer} involves adding or altering the model parameters responsible for the undesirable output.
Meta-learning methods~\citep{de2021editing, mitchell2021fast} use a hypernetwork to learn the necessary adjustments for editing LLMs.
In-context editing methods (ICE)~\citep{mitchell2022memory, madaan2022memory, zhong2023mquake, zheng2023can} demonstrate significant potential, enabling the editing of LLMs by prompting them with edited facts and retrieving editing demonstrations from the edit memory.

\paragraph{Decoding Strategy.} 
Recent work modifies various decoding strategies to enhance different alignments by altering the logits of the original tokens during generation.
CD~\citep{li-etal-2023-contrastive} compares powerful expert language models with weaker amateur language models to enhance fluency and coherence.
DoLa~\citep{chuang2023dola} contrasts mature layers with premature layers, while ICD~\citep{zhang2023alleviating} compares with models injected with hallucinations, aiming to enhance the factual accuracy of the model.
DeCK~\citep{bi2024decoding} enhances ICE by highlighting the output probability increment of new knowledge in contrast to the parametric knowledge.
Unlike the aforementioned decoding methods, \ours proposed in this paper only needs to adjust key tokens to enhance KE and mitigate factual hallucinations in LLMs.

\section{Conclusion}

In this work, we propose a new KE framework, \ours, to enhance ICE. \ours focuses on the crucial tokens that are mostly related to knowledge during the generation, biasing their logits by matching the knowledge entities. 
This design effectively reduces the potential risk of introducing fundamental errors in the logical coherence of the entire inference statement. 
Experimental results show that \ours significantly improves the editing success rate of ICE and outperforms the current best decoding methods. 
Furthermore, the latency of \ours is at most 1.01 times that of the baseline, meaning \ours not only enhances ICE but can also be widely applied with negligible cost.

\section*{Limitations}

We mainly evaluate the KE methods on the \textsc{LLaMA2-7B-chat}, \textsc{LLaMA2-13B-chat}, and \Mistral. The efficacy of these methods on other LLMs remains less explored. Additionally, although \ours is expected to be easily deployable on any ICE method to enhance KE performance, we currently evaluate \ours on the representative IKE and MeLLo, lacking broader validation. We leave the evaluation on other models and ICE methods for future work.

\section*{Ethical Considerations}
Ethical considerations are of utmost importance in our research endeavors.  In this paper, we conscientiously adhere to ethical principles by exclusively utilizing open-source datasets and employing models that are either open-source or widely recognized in the scientific community.  Moreover, counterfactual public datasets were used in knowledge editing to measure knowledge updates. Our proposed method is designed to ensure that the model does not produce any harmful or misleading information. We are committed to upholding ethical standards throughout the research process, prioritizing transparency, and promoting the responsible use of technology for the betterment of society.

\bibliography{custom}

\begin{thebibliography}{35}
\providecommand{\natexlab}[1]{#1}

\bibitem[{Bi et~al.(2024{\natexlab{a}})Bi, Liu, Mei, Wang, Ji, and Cheng}]{bi2024decoding}
Baolong Bi, Shenghua Liu, Lingrui Mei, Yiwei Wang, Pengliang Ji, and Xueqi Cheng. 2024{\natexlab{a}}.
\newblock \href {https://arxiv.org/abs/2405.11613} {Decoding by contrasting knowledge: Enhancing llms' confidence on edited facts}.
\newblock \emph{Preprint}, arXiv:2405.11613.

\bibitem[{Bi et~al.(2024{\natexlab{b}})Bi, Liu, Wang, Mei, and Cheng}]{bi2024factuality}
Baolong Bi, Shenghua Liu, Yiwei Wang, Lingrui Mei, and Xueqi Cheng. 2024{\natexlab{b}}.
\newblock Is factuality decoding a free lunch for llms? evaluation on knowledge editing benchmark.
\newblock \emph{arXiv preprint arXiv:2404.00216}.

\bibitem[{Chen and Shu(2023)}]{chen2023combating}
Canyu Chen and Kai Shu. 2023.
\newblock Combating misinformation in the age of llms: Opportunities and challenges.
\newblock \emph{arXiv preprint arXiv:2311.05656}.

\bibitem[{Chuang et~al.(2023)Chuang, Xie, Luo, Kim, Glass, and He}]{chuang2023dola}
Yung-Sung Chuang, Yujia Xie, Hongyin Luo, Yoon Kim, James Glass, and Pengcheng He. 2023.
\newblock Dola: Decoding by contrasting layers improves factuality in large language models.
\newblock \emph{arXiv preprint arXiv:2309.03883}.

\bibitem[{Cohen et~al.(2024)Cohen, Biran, Yoran, Globerson, and Geva}]{cohen2024evaluating}
Roi Cohen, Eden Biran, Ori Yoran, Amir Globerson, and Mor Geva. 2024.
\newblock Evaluating the ripple effects of knowledge editing in language models.
\newblock \emph{Transactions of the Association for Computational Linguistics}, 12:283--298.

\bibitem[{De~Cao et~al.(2021)De~Cao, Aziz, and Titov}]{de2021editing}
Nicola De~Cao, Wilker Aziz, and Ivan Titov. 2021.
\newblock Editing factual knowledge in language models.
\newblock \emph{arXiv preprint arXiv:2104.08164}.

\bibitem[{Huang et~al.(2023{\natexlab{a}})Huang, Yu, Ma, Zhong, Feng, Wang, Chen, Peng, Feng, Qin, and Liu}]{huang2023survey}
Lei Huang, Weijiang Yu, Weitao Ma, Weihong Zhong, Zhangyin Feng, Haotian Wang, Qianglong Chen, Weihua Peng, Xiaocheng Feng, Bing Qin, and Ting Liu. 2023{\natexlab{a}}.
\newblock \href {https://arxiv.org/abs/2311.05232} {A survey on hallucination in large language models: Principles, taxonomy, challenges, and open questions}.
\newblock \emph{Preprint}, arXiv:2311.05232.

\bibitem[{Huang et~al.(2023{\natexlab{b}})Huang, Shen, Zhang, Zhou, Rong, and Xiong}]{huang2023transformer}
Zeyu Huang, Yikang Shen, Xiaofeng Zhang, Jie Zhou, Wenge Rong, and Zhang Xiong. 2023{\natexlab{b}}.
\newblock Transformer-patcher: One mistake worth one neuron.
\newblock \emph{arXiv preprint arXiv:2301.09785}.

\bibitem[{Jiang et~al.(2023)Jiang, Sablayrolles, Mensch, Bamford, Chaplot, Casas, Bressand, Lengyel, Lample, Saulnier et~al.}]{jiang2023mistral}
Albert~Q Jiang, Alexandre Sablayrolles, Arthur Mensch, Chris Bamford, Devendra~Singh Chaplot, Diego de~las Casas, Florian Bressand, Gianna Lengyel, Guillaume Lample, Lucile Saulnier, et~al. 2023.
\newblock Mistral 7b.
\newblock \emph{arXiv preprint arXiv:2310.06825}.

\bibitem[{Jiang et~al.(2024)Jiang, Wu, Xiong, Ruan, Ding, Guo, Wen, Zhou, and Deng}]{jiang2024hummer}
Li~Jiang, Yusen Wu, Junwu Xiong, Jingqing Ruan, Yichuan Ding, Qingpei Guo, Zujie Wen, Jun Zhou, and Xiaotie Deng. 2024.
\newblock Hummer: Towards limited competitive preference dataset.
\newblock \emph{arXiv preprint arXiv:2405.11647}.

\bibitem[{Li et~al.(2023)Li, Holtzman, Fried, Liang, Eisner, Hashimoto, Zettlemoyer, and Lewis}]{li-etal-2023-contrastive}
Xiang~Lisa Li, Ari Holtzman, Daniel Fried, Percy Liang, Jason Eisner, Tatsunori Hashimoto, Luke Zettlemoyer, and Mike Lewis. 2023.
\newblock \href {https://doi.org/10.18653/v1/2023.acl-long.687} {Contrastive decoding: Open-ended text generation as optimization}.
\newblock In \emph{Proceedings of the 61st Annual Meeting of the Association for Computational Linguistics (Volume 1: Long Papers)}, pages 12286--12312, Toronto, Canada. Association for Computational Linguistics.

\bibitem[{Madaan et~al.(2022)Madaan, Tandon, Clark, and Yang}]{madaan2022memory}
Aman Madaan, Niket Tandon, Peter Clark, and Yiming Yang. 2022.
\newblock Memory-assisted prompt editing to improve gpt-3 after deployment.
\newblock \emph{arXiv preprint arXiv:2201.06009}.

\bibitem[{Meng et~al.(2022{\natexlab{a}})Meng, Bau, Andonian, and Belinkov}]{meng2022locating}
Kevin Meng, David Bau, Alex Andonian, and Yonatan Belinkov. 2022{\natexlab{a}}.
\newblock Locating and editing factual associations in gpt.
\newblock \emph{Advances in Neural Information Processing Systems}, 35:17359--17372.

\bibitem[{Meng et~al.(2022{\natexlab{b}})Meng, Sharma, Andonian, Belinkov, and Bau}]{meng2022mass}
Kevin Meng, Arnab~Sen Sharma, Alex Andonian, Yonatan Belinkov, and David Bau. 2022{\natexlab{b}}.
\newblock Mass-editing memory in a transformer.
\newblock \emph{arXiv preprint arXiv:2210.07229}.

\bibitem[{Mitchell et~al.(2021)Mitchell, Lin, Bosselut, Finn, and Manning}]{mitchell2021fast}
Eric Mitchell, Charles Lin, Antoine Bosselut, Chelsea Finn, and Christopher~D Manning. 2021.
\newblock Fast model editing at scale.
\newblock \emph{arXiv preprint arXiv:2110.11309}.

\bibitem[{Mitchell et~al.(2022)Mitchell, Lin, Bosselut, Manning, and Finn}]{mitchell2022memory}
Eric Mitchell, Charles Lin, Antoine Bosselut, Christopher~D Manning, and Chelsea Finn. 2022.
\newblock Memory-based model editing at scale.
\newblock In \emph{International Conference on Machine Learning}, pages 15817--15831. PMLR.

\bibitem[{Niwattanakul et~al.(2013)Niwattanakul, Singthongchai, Naenudorn, and Wanapu}]{niwattanakul2013using}
Suphakit Niwattanakul, Jatsada Singthongchai, Ekkachai Naenudorn, and Supachanun Wanapu. 2013.
\newblock Using of jaccard coefficient for keywords similarity.
\newblock In \emph{Proceedings of the international multiconference of engineers and computer scientists}, volume~1, pages 380--384.

\bibitem[{OpenAI(2022)}]{chatgpt}
OpenAI. 2022.
\newblock \href {https://openai.com/blog/chatgpt} {large-scale generative pre-training model for conversation}.
\newblock \emph{OpenAI blog}.

\bibitem[{OpenAI(2023)}]{openai2023gpt4}
OpenAI. 2023.
\newblock \href {https://arxiv.org/abs/2303.08774} {Gpt-4 technical report}.
\newblock \emph{Preprint}, arXiv:2303.08774.

\bibitem[{Shi et~al.(2024)Shi, Tan, Wu, Zhong, Zhou, and Liu}]{shi2024retrieval}
Yucheng Shi, Qiaoyu Tan, Xuansheng Wu, Shaochen Zhong, Kaixiong Zhou, and Ninghao Liu. 2024.
\newblock Retrieval-enhanced knowledge editing for multi-hop question answering in language models.
\newblock \emph{arXiv preprint arXiv:2403.19631}.

\bibitem[{Sinitsin et~al.(2020)Sinitsin, Plokhotnyuk, Pyrkin, Popov, and Babenko}]{sinitsin2020editable}
Anton Sinitsin, Vsevolod Plokhotnyuk, Dmitriy Pyrkin, Sergei Popov, and Artem Babenko. 2020.
\newblock Editable neural networks.
\newblock \emph{arXiv preprint arXiv:2004.00345}.

\bibitem[{Song et~al.(2024)Song, Yuan, and Yang}]{song2024fmint}
Zezheng Song, Jiaxin Yuan, and Haizhao Yang. 2024.
\newblock Fmint: Bridging human designed and data pretrained models for differential equation foundation model.
\newblock \emph{arXiv preprint arXiv:2404.14688}.

\bibitem[{Tonmoy et~al.(2024)Tonmoy, Zaman, Jain, Rani, Rawte, Chadha, and Das}]{tonmoy2024comprehensive}
S.~M Towhidul~Islam Tonmoy, S~M~Mehedi Zaman, Vinija Jain, Anku Rani, Vipula Rawte, Aman Chadha, and Amitava Das. 2024.
\newblock \href {https://arxiv.org/abs/2401.01313} {A comprehensive survey of hallucination mitigation techniques in large language models}.
\newblock \emph{Preprint}, arXiv:2401.01313.

\bibitem[{Touvron et~al.(2023{\natexlab{a}})Touvron, Lavril, Izacard, Martinet, Lachaux, Lacroix, Rozi{\`{e}}re, Goyal, Hambro, Azhar, Rodriguez, Joulin, Grave, and Lample}]{DBLP:journals/corr/abs-2302-13971}
Hugo Touvron, Thibaut Lavril, Gautier Izacard, Xavier Martinet, Marie{-}Anne Lachaux, Timoth{\'{e}}e Lacroix, Baptiste Rozi{\`{e}}re, Naman Goyal, Eric Hambro, Faisal Azhar, Aur{\'{e}}lien Rodriguez, Armand Joulin, Edouard Grave, and Guillaume Lample. 2023{\natexlab{a}}.
\newblock \href {https://doi.org/10.48550/ARXIV.2302.13971} {Llama: Open and efficient foundation language models}.
\newblock \emph{CoRR}, abs/2302.13971.

\bibitem[{Touvron et~al.(2023{\natexlab{b}})Touvron, Martin, Stone, Albert, Almahairi, Babaei, Bashlykov, Batra, Bhargava, Bhosale, Bikel, Blecher, Ferrer, Chen, Cucurull, Esiobu, Fernandes, Fu, Fu, Fuller, Gao, Goswami, Goyal, Hartshorn, Hosseini, Hou, Inan, Kardas, Kerkez, Khabsa, Kloumann, Korenev, Koura, Lachaux, Lavril, Lee, Liskovich, Lu, Mao, Martinet, Mihaylov, Mishra, Molybog, Nie, Poulton, Reizenstein, Rungta, Saladi, Schelten, Silva, Smith, Subramanian, Tan, Tang, Taylor, Williams, Kuan, Xu, Yan, Zarov, Zhang, Fan, Kambadur, Narang, Rodriguez, Stojnic, Edunov, and Scialom}]{touvron2023llama}
Hugo Touvron, Louis Martin, Kevin Stone, Peter Albert, Amjad Almahairi, Yasmine Babaei, Nikolay Bashlykov, Soumya Batra, Prajjwal Bhargava, Shruti Bhosale, Dan Bikel, Lukas Blecher, Cristian~Canton Ferrer, Moya Chen, Guillem Cucurull, David Esiobu, Jude Fernandes, Jeremy Fu, Wenyin Fu, Brian Fuller, Cynthia Gao, Vedanuj Goswami, Naman Goyal, Anthony Hartshorn, Saghar Hosseini, Rui Hou, Hakan Inan, Marcin Kardas, Viktor Kerkez, Madian Khabsa, Isabel Kloumann, Artem Korenev, Punit~Singh Koura, Marie-Anne Lachaux, Thibaut Lavril, Jenya Lee, Diana Liskovich, Yinghai Lu, Yuning Mao, Xavier Martinet, Todor Mihaylov, Pushkar Mishra, Igor Molybog, Yixin Nie, Andrew Poulton, Jeremy Reizenstein, Rashi Rungta, Kalyan Saladi, Alan Schelten, Ruan Silva, Eric~Michael Smith, Ranjan Subramanian, Xiaoqing~Ellen Tan, Binh Tang, Ross Taylor, Adina Williams, Jian~Xiang Kuan, Puxin Xu, Zheng Yan, Iliyan Zarov, Yuchen Zhang, Angela Fan, Melanie Kambadur, Sharan Narang, Aurelien Rodriguez, Robert Stojnic, Sergey Edunov, and Thomas
  Scialom. 2023{\natexlab{b}}.
\newblock \href {https://arxiv.org/abs/2307.09288} {Llama 2: Open foundation and fine-tuned chat models}.
\newblock \emph{Preprint}, arXiv:2307.09288.

\bibitem[{Wang et~al.(2023{\natexlab{a}})Wang, Liu, Yue, Tang, Zhang, Jiayang, Yao, Gao, Hu, Qi et~al.}]{wang2023survey}
Cunxiang Wang, Xiaoze Liu, Yuanhao Yue, Xiangru Tang, Tianhang Zhang, Cheng Jiayang, Yunzhi Yao, Wenyang Gao, Xuming Hu, Zehan Qi, et~al. 2023{\natexlab{a}}.
\newblock Survey on factuality in large language models: Knowledge, retrieval and domain-specificity.
\newblock \emph{arXiv preprint arXiv:2310.07521}.

\bibitem[{Wang et~al.(2023{\natexlab{b}})Wang, Zhang, Xie, Yao, Tian, Wang, Xi, Cheng, Liu, Zheng et~al.}]{wang2023easyedit}
Peng Wang, Ningyu Zhang, Xin Xie, Yunzhi Yao, Bozhong Tian, Mengru Wang, Zekun Xi, Siyuan Cheng, Kangwei Liu, Guozhou Zheng, et~al. 2023{\natexlab{b}}.
\newblock Easyedit: An easy-to-use knowledge editing framework for large language models.
\newblock \emph{arXiv preprint arXiv:2308.07269}.

\bibitem[{Wang et~al.(2024)Wang, Chen, Peng, and Chang}]{wang2024deepedit}
Yiwei Wang, Muhao Chen, Nanyun Peng, and Kai-Wei Chang. 2024.
\newblock Deepedit: Knowledge editing as decoding with constraints.
\newblock \emph{arXiv preprint arXiv:2401.10471}.

\bibitem[{Wei et~al.(2022)Wei, Wang, Schuurmans, Bosma, Xia, Chi, Le, Zhou et~al.}]{wei2022chain}
Jason Wei, Xuezhi Wang, Dale Schuurmans, Maarten Bosma, Fei Xia, Ed~Chi, Quoc~V Le, Denny Zhou, et~al. 2022.
\newblock Chain-of-thought prompting elicits reasoning in large language models.
\newblock \emph{Advances in neural information processing systems}, 35:24824--24837.

\bibitem[{Yao et~al.(2023)Yao, Wang, Tian, Cheng, Li, Deng, Chen, and Zhang}]{yao2023editing}
Yunzhi Yao, Peng Wang, Bozhong Tian, Siyuan Cheng, Zhoubo Li, Shumin Deng, Huajun Chen, and Ningyu Zhang. 2023.
\newblock Editing large language models: Problems, methods, and opportunities.
\newblock \emph{arXiv preprint arXiv:2305.13172}.

\bibitem[{Zhang et~al.(2023{\natexlab{a}})Zhang, Cui, Bi, and Shi}]{zhang2023alleviating}
Yue Zhang, Leyang Cui, Wei Bi, and Shuming Shi. 2023{\natexlab{a}}.
\newblock Alleviating hallucinations of large language models through induced hallucinations.
\newblock \emph{arXiv preprint arXiv:2312.15710}.

\bibitem[{Zhang et~al.(2023{\natexlab{b}})Zhang, Li, Cui, Cai, Liu, Fu, Huang, Zhao, Zhang, Chen, Wang, Luu, Bi, Shi, and Shi}]{zhang2023sirens}
Yue Zhang, Yafu Li, Leyang Cui, Deng Cai, Lemao Liu, Tingchen Fu, Xinting Huang, Enbo Zhao, Yu~Zhang, Yulong Chen, Longyue Wang, Anh~Tuan Luu, Wei Bi, Freda Shi, and Shuming Shi. 2023{\natexlab{b}}.
\newblock \href {https://arxiv.org/abs/2309.01219} {Siren's song in the ai ocean: A survey on hallucination in large language models}.
\newblock \emph{Preprint}, arXiv:2309.01219.

\bibitem[{Zheng et~al.(2023)Zheng, Li, Dong, Fan, Wu, Xu, and Chang}]{zheng2023can}
Ce~Zheng, Lei Li, Qingxiu Dong, Yuxuan Fan, Zhiyong Wu, Jingjing Xu, and Baobao Chang. 2023.
\newblock Can we edit factual knowledge by in-context learning?
\newblock \emph{arXiv preprint arXiv:2305.12740}.

\bibitem[{Zhong et~al.(2023)Zhong, Wu, Manning, Potts, and Chen}]{zhong2023mquake}
Zexuan Zhong, Zhengxuan Wu, Christopher~D Manning, Christopher Potts, and Danqi Chen. 2023.
\newblock Mquake: Assessing knowledge editing in language models via multi-hop questions.
\newblock \emph{arXiv preprint arXiv:2305.14795}.

\bibitem[{Zhu et~al.(2020)Zhu, Rawat, Zaheer, Bhojanapalli, Li, Yu, and Kumar}]{zhu2020modifying}
Chen Zhu, Ankit~Singh Rawat, Manzil Zaheer, Srinadh Bhojanapalli, Daliang Li, Felix Yu, and Sanjiv Kumar. 2020.
\newblock Modifying memories in transformer models.
\newblock \emph{arXiv preprint arXiv:2012.00363}.

\end{thebibliography}

\appendix
% \clearpage

\begin{figure*}[t!]
    \centering
    \includegraphics[width=0.9\linewidth]{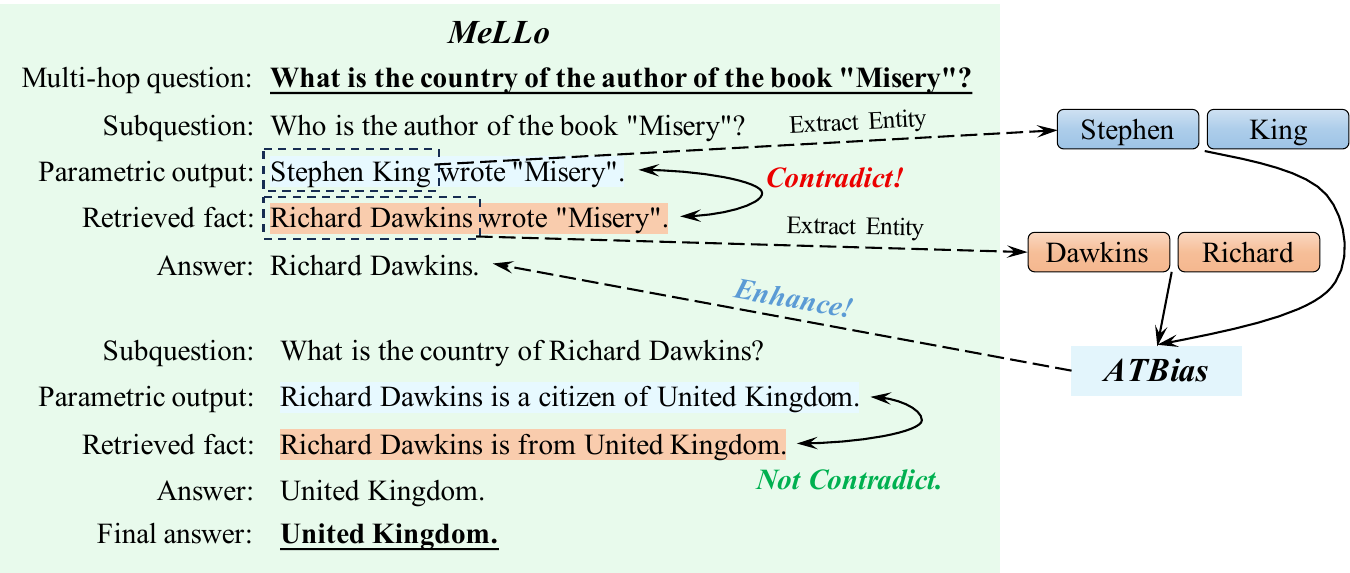}
    % \vspace{-8mm}
    \caption{An illustration of \ours's easy deployment on MeLLo.}
    \label{fig:mello}
    \vspace{-4mm}
\end{figure*}

\section{How Can \ours Be Easily Deployed on MeLLo?}
\label{sec:mello_example}
Many advanced ICE methods~\citep{zhong2023mquake, wang2024deepedit, shi2024retrieval} inherently possess parametric knowledge, so \ours does not need to induce LLMs to preprocess it offline.
Table \ref{fig:mello} demonstrates how MeLLo~\citep{zhong2023mquake} can easily deploy \ours without additional inference, directly extracting the required entities from the parametric output of subquestion responses.
This means that knowledge entities can be extracted online and fed into \ours when using MeLLo. Thus, \ours enhances ICE during the decoding stage with just a single inference step.

\section{Additional Ablation Study of \ours}
\label{sec:add_ablation}

We conduct following additional ablation study experiments using the ICE method IKE~\citep{zheng2023can} with \llamaa and \llamab on the \textsc{MQuAKE-3k} datasets.

\subsection{N-gram Decomposition}
\begin{figure}[h]
    \centering
    \includegraphics[width=\linewidth]{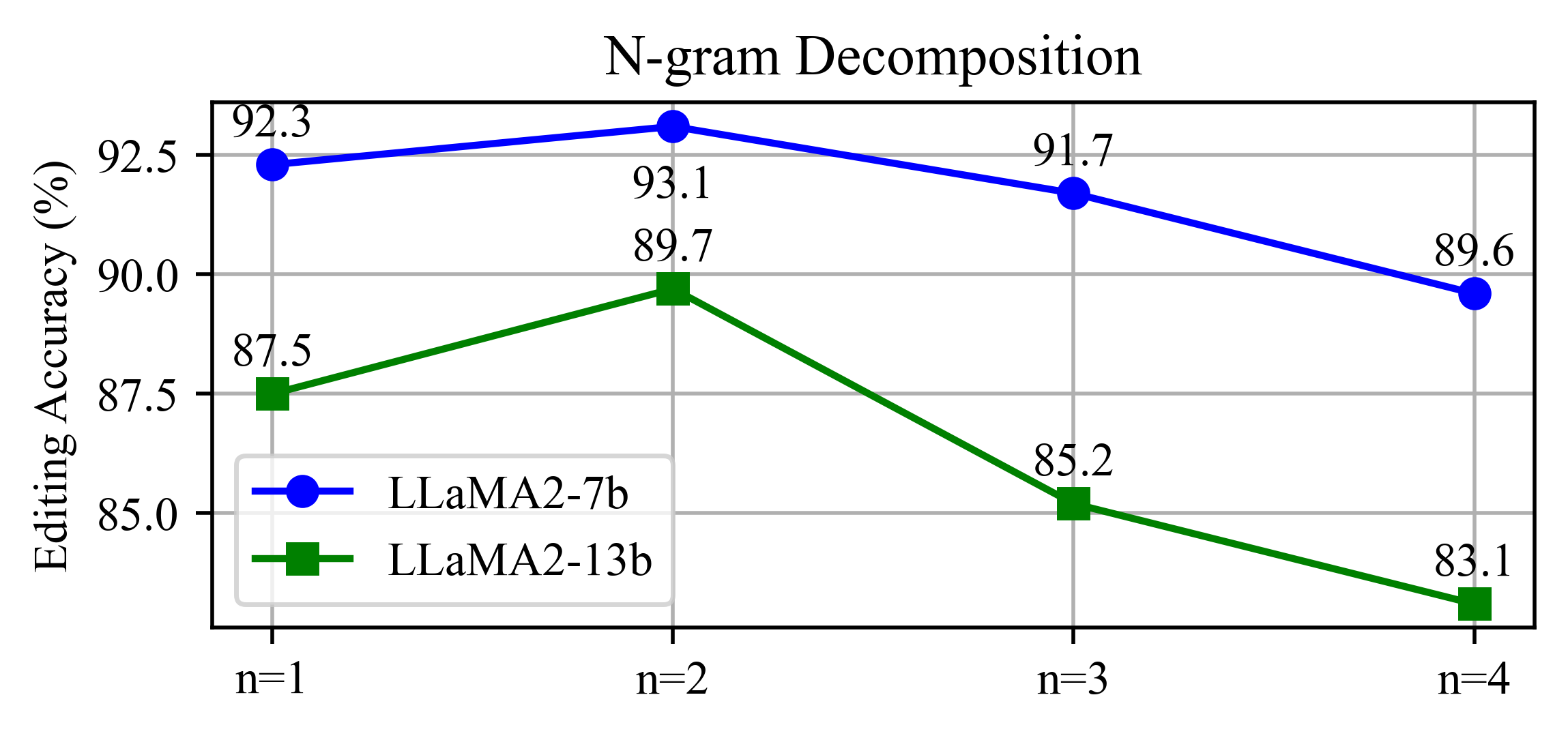}
    \vspace{-8mm}
    \caption{Ablation study results of the gram n for n-gram decomposition process.}
    \label{fig:ablation_1}
    \vspace{-6mm}
\end{figure}

The N-gram decomposition is a prerequisite for calculating the similarity between the knowledge entities and filtered tokens (Section \ref{sec:similarity}). 
Table \ref{fig:ablation_1} presents the ablation study results for various values of gram n during this process. Both excessively high and low decomposition precision can diminish the matching effectiveness, with $n=2$ yielding the best editing performance.

\subsection{Probabilistic Constraint of Filter}

\begin{figure}[h]
    \centering
    \includegraphics[width=\linewidth]{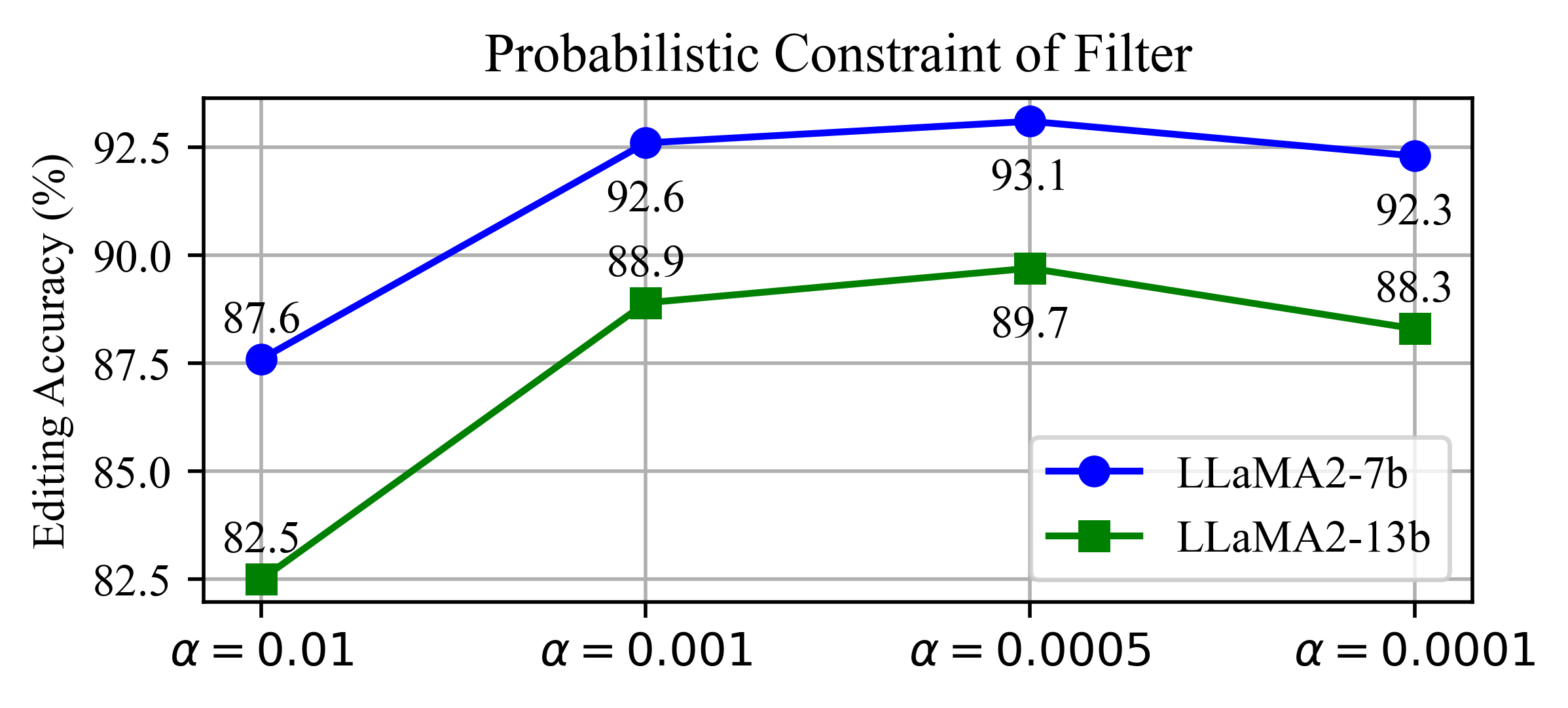}
    \vspace{-8mm}
    \caption{Ablation study results of the probabilistic constraint $\alpha$ of filter.}
    \label{fig:ablation_2}
    \vspace{-6mm}
\end{figure}

The probabilistic constraint of \ours's filter (Section \ref{sec:filter}) that represented in Equation \ref{eq:prob} is subjected to an ablation study on the parameter $\alpha$.
The results of this study are shown in Table \ref{fig:ablation_2}, indicating that $\alpha=0.0005$ yields the best editing performance. 
The fact that smaller $\alpha$ values yield better performance further indicates the strictness of our filtering process, effectively preventing interference from unreasonable tokens.

\subsection{Ranking Constraint of Filter}

\begin{figure}[h]
    \centering
    \includegraphics[width=\linewidth]{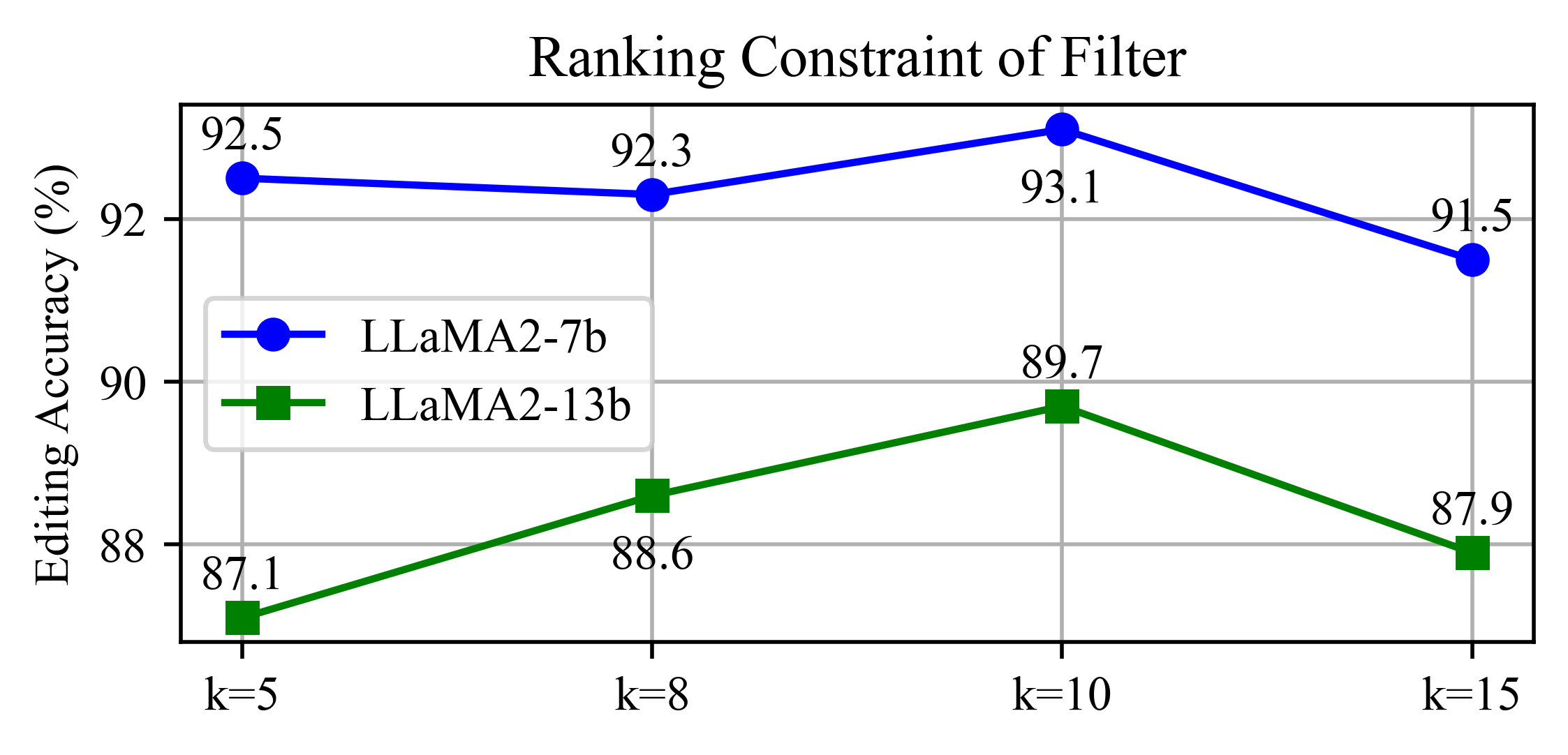}
    \vspace{-8mm}
    \caption{Ablation study results of the ranking constraint $k$ of filter.}
    \label{fig:ablation_3}
    \vspace{-6mm}
\end{figure}

The ablation study results of ranking constraint (Equation \ref{eq:rank}) are illustrated in Table \ref{fig:ablation_3}, showing that $k=10$ yields the best editing performance.

\begin{table*}[ht]
    \centering
    \small
    \noindent\fbox{%
    \begin{minipage}{2.0\columnwidth} 
    \tt 
    \textcolor{gray}{$[$3 in-context demonstrations abbreviated$]$} \\ \\
        Question: \textbf{\underline{What is the capital city of the country of citizenship of Ivanka Trump's spouse?}}\\
        Edit Knowledge: \orangetext{Jared Kushner is a citizen of Canada}.\\
        Thoughts: Ivanka Trump's spouse is Jared Kushner. \orangetext{Jared Kushner is a citizen of Canada}. The capital city of Canada is Ottawa.\\
        Answer: \textbf{\underline{Ottawa}}\\ \\
        Question: \textbf{\underline{Which continent is the country where the director of "My House Husband: Ikaw Na!" }}\\
        \textbf{\underline{was educated located in?}}\\
        Edit Knowledge: \orangetext{Irene Villamor was educated in New York University}.\\
        Thoughts: The director of "My House Husband: Ikaw Na!" is Jose Javier Reyes. \orangetext{Jose Javier Reyes was educated in New York University}. De La Salle University is located in United States of America. United States of America is located in the continent if North America.\\
        Answer: \textbf{\underline{North America}}
    \end{minipage}
    }

    \caption{An illustration of the COT based IKE solving two simplified examples. The orange parts are facts retrieved by the retriever.}
    \label{tab:prompts_ike}
\end{table*}

\definecolor{maskcolor}{HTML}{6e9feb}

\begin{table*}[ht]
    \centering
    \small
    \noindent\fbox{%
    \begin{minipage}{2.0\columnwidth} 
    \tt 
    \textcolor{gray}{$[$2 in-context demonstrations abbreviated$]$} \\ \\
        Question: \textbf{\underline{What is the capital city of the country of citizenship of Ivanka Trump's spouse?}}\\
        Subquestion: Who is Ivanka Trump's spouse?\\
        Generated answer:\bluetext{Ivanka Trump's spouse is Jared Kushner.}\\
        Retrieved fact:\orangetext{David Cameron is married to Samantha Cameron.}\\
        Retrieved fact does not contradict to generated answer.\\
        Intermediate answer: \bluetext{Jared Kushner}\\
        Subquestion: What is the country of citizenship of Jared Kushner?\\
        Generated answer:\bluetext{The country of citizenship of Jared Kushner is United States.}\\
        Retrieved fact:\orangetext{Jared Kushner is a citizen of Canada.}\\
        Retrieved fact contradicts to generated answer.\\
        Intermediate answer: \orangetext{Canada}\\
        Subquestion: What is the capital city of Canada?\\
        Generated answer:\bluetext{The capital city of Canada is Ottawa.}\\
        Retrieved fact:\orangetext{The capital city of United States is Seattle.}\\
        Retrieved fact does not contradict to generated answer, so the intermediate answer.\\
        Intermediate answer: \bluetext{Ottawa}\\
        Final answer: \textbf{\underline{Ottawa}}
    \end{minipage}
    }

    \caption{A step-by-step illustration of MeLLo solving one simplified example. Blue parts are generated by the language model, and orange parts are facts retrieved by the retriever.}
    \label{tab:prompts_mello}
\end{table*}

\subsection{Bias Coefficient of Knowledge}

We adjust the logits of tokens matching with the new and parametric knowledge entities (Section \ref{eq:adjustment1}) with the bias coefficients $\lambda_n$ (Equation \ref{eq:adjustment2}) and $\lambda_p$.
The ablation study results of $\lambda_n$ and  $\lambda_p$ are shown in Table \ref{tab:ablation-new} and \ref{tab:ablation-para}, respectively.
\ours achieves the best performance when $\lambda_n=25$.

\begin{table}[ht]
\centering
\small
\renewcommand\arraystretch{1.3}
\setlength{\tabcolsep}{8pt}
\begin{tabular}{lccc} 
\toprule
 \ \textbf{Model} & \textbf{$\lambda_{n}=20$}  & \textbf{$\lambda_{n}=25$} & \textbf{$\lambda_{n}=30$}\\ \midrule 
 \textsc{LLaMA2-7B} & 90.5 & \bf 93.1 &  92.7 \\ 
 \textsc{LLaMA2-13B} & 86.6 & \bf 89.7 & 88.9 \\ 
\bottomrule
\end{tabular}
\caption{Ablation study results of the bias coefficient of new knowledge $\lambda_n$.} 
\label{tab:ablation-new} 
\vspace{-10pt}
\end{table}

\ours achieves the best performance when $\lambda_p=1$.
An $\lambda_p$ value of $0$ means that the logits of tokens matching with parametric knowledge entities are not reduced, and the results indicate that this leads to a decline in performance.
Optimal performance is achieved with smaller values of $\lambda_p$ because excessively large 
$\lambda_p$ values may cause the logits of tokens incorrectly matching old knowledge entities to decrease too much, adversely affecting editing performance.

\begin{table}[ht]
\centering
\small
\renewcommand\arraystretch{1.3}
\setlength{\tabcolsep}{8pt}
\begin{tabular}{lccc} 
\toprule
 \ \textbf{Model} & \textbf{$\lambda_{p}=0$}  & \textbf{$\lambda_{p}=1$} & \textbf{$\lambda_{p}=2$}\\ \midrule 
 \textsc{LLaMA2-7B} & 85.9 & \bf 93.1 &  88.6 \\ 
 \textsc{LLaMA2-13B} & 70.2 & \bf 89.7 & 83.2 \\ 
\bottomrule
\end{tabular}
\caption{Ablation study results of the bias coefficient of parametric knowledge $\lambda_p$.} 
\label{tab:ablation-para} 
\vspace{-10pt}
\end{table}

\section{Prompts of ICE for Experiments}
\label{sec:prompts}
The prompt we used in IKE~\citep{zheng2023can} is shown in \ref{tab:prompts_ike}, and the prompt we used in MeLLo is shown in \ref{tab:prompts_mello}.
Based on the provided contextual demonstrations, LLMs can be guided to perform the corresponding ICE methods. 
\ours can enhance these ICE methods without modifying any prompts.

\end{document}